\documentclass[10pt,twocolumn,letterpaper]{article}

\usepackage{iccv}              

\definecolor{iccvblue}{rgb}{0.21,0.49,0.74}
\usepackage[pagebackref,breaklinks,colorlinks,allcolors=iccvblue]{hyperref}

\usepackage{multirow}
\usepackage{booktabs}

\def\paperID{8131} 
\def\confName{ICCV}
\def\confYear{2025}

\title{MotionAgent: Fine-grained Controllable Video Generation\\ via Motion Field Agent}

\author{
  Xinyao Liao$^{1,2}$\qquad
  Xianfang Zeng$^{2}$\footnotemark[1]\qquad
  Liao Wang$^{2}$\qquad 
  Gang Yu$^{2}$\footnotemark[2]\qquad 
  Guosheng Lin$^{1}$\footnotemark[2]\qquad
  Chi Zhang$^{3}$\\
  $^{1}$Nanyang Technological University \qquad
  $^{2}$StepFun \qquad
  $^{3}$Westlake University\\[1ex]
  \url{https://github.com/leoisufa/MotionAgent}
}

\begin{document}
\maketitle
\renewcommand{\thefootnote}{\fnsymbol{footnote}}
\footnotetext[1]{Xianfang Zeng is the project leader.}
\footnotetext[2]{Corresponding authors: skicy@outlook.com, gslin@ntu.edu.sg}
\renewcommand{\thefootnote}{\arabic{footnote}}
\begin{abstract}
We propose MotionAgent, enabling fine-grained motion control for text-guided image-to-video generation. The key technique is the motion field agent that converts motion information in text prompts into explicit motion fields, providing flexible and precise motion guidance. Specifically, the agent extracts the object movement and camera motion described in the text, and converts them into object trajectories and camera extrinsics, respectively. An analytical optical flow composition module integrates these motion representations in 3D space and projects them into a unified optical flow. An optical flow adapter takes the flow to control the base image-to-video diffusion model for generating fine-grained controlled videos. After that, an optional rethinking step can be adopted to ensure the generated video is aligned well with motion information in the prompt. The significant improvement in the Video-Text Camera Motion metrics on VBench indicates that our method achieves precise control over camera motion. We further construct a subset of VBench to evaluate the alignment of motion information in the text and the generated video, outperforming other advanced models on motion generation accuracy.
\end{abstract}    
\section{Introduction}
\label{sec:introduction}

\begin{figure}[t]
\centering
\includegraphics[width=0.99\columnwidth]{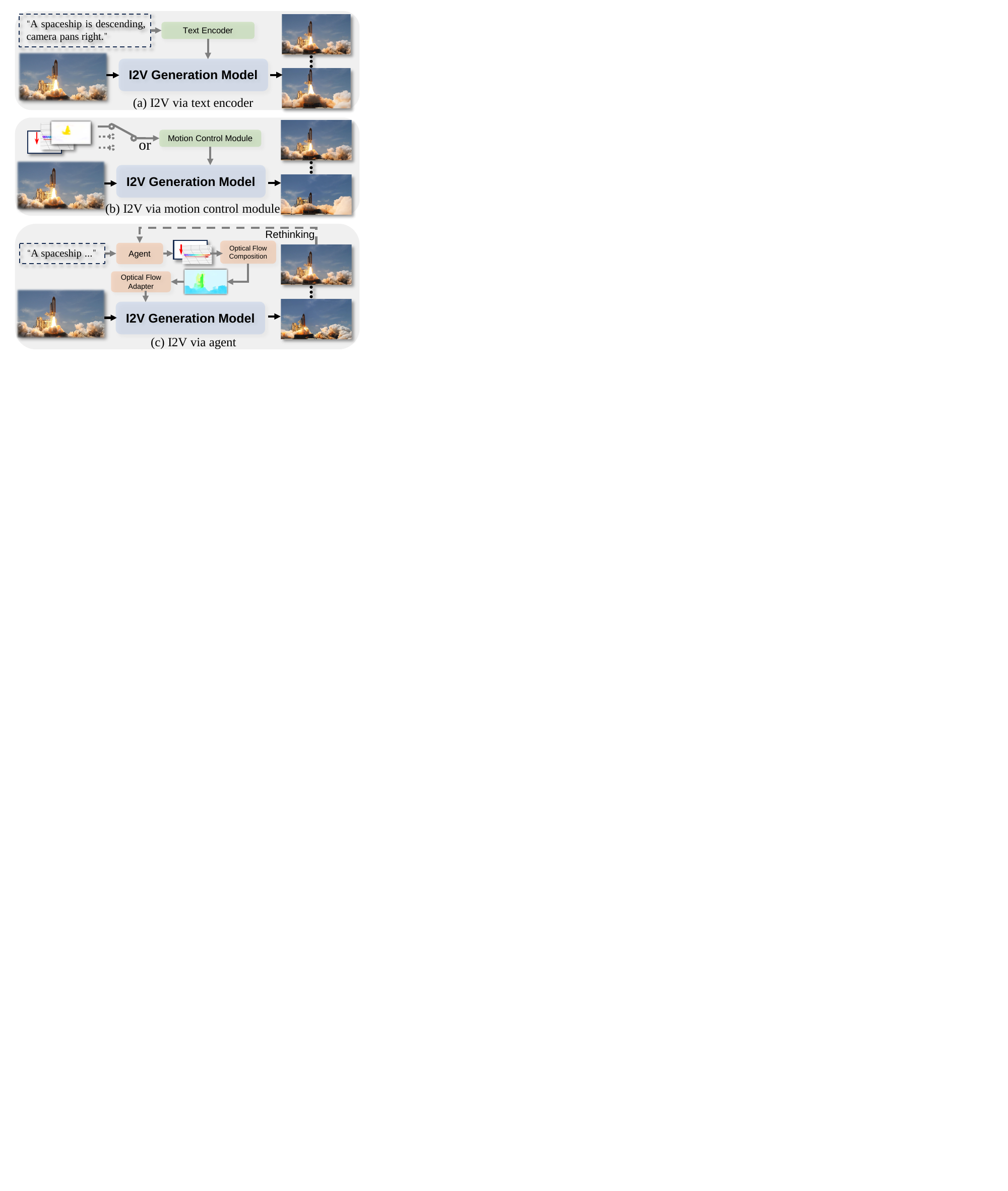}
\caption{Different frameworks of I2V generation models. (a) Controllable I2V generation via text encoder. (b) Controllable I2V generation via special control module. (c) Our method, controllable I2V generation via motion field agent.}
\label{fig:framework}
\end{figure}

Recently, image-to-video (I2V) generation models \cite{chen2023videocrafter1,dai2023animateanything,zhang2023i2vgen,guo2024i2v,ren2024consisti2v,xing2025dynamicrafter,zhang2024moonshot,ma2024cinemo,blattmann2023stable,ma2024follow,jin2024pyramidal,yang2024cogvideox} develop rapidly. These models bring images to life, making the visual content more dynamic and vivid. Compared to text-to-video (T2V) generation models \cite{ho2022video,ho2022imagen,singer2022make,ge2023preserve,mei2023vidm,he2022latent,an2023latent,blattmann2023align,guo2023animatediff}, I2V models use an image as the reference, which further constrains the content and reduces the uncertainty of the generated video. Most existing I2V generation models already achieve high-quality video generation, allowing them to preserve the visual details of the input image while generating stable results. However, precise control over the video using only text input remains a field worth exploring.

Some I2V generation models \cite{guo2024i2v,blattmann2023stable} randomly make the input image dynamic, resulting in an uncontrollable generated result. Other I2V models \cite{xing2025dynamicrafter,zhang2023i2vgen,ren2024consisti2v,dai2023animateanything,zhang2024moonshot,ma2024cinemo,ma2024follow} support simple control through text input. As shown in Figure \ref{fig:framework}(a), they adopt a text encoder to inject control information into visual features. They can perform overall control, but it is difficult to achieve fine-grained control over each element of the video. Additionally, training an I2V generation model with high alignment between motion information in the text and the video relies heavily on the quality of training data \cite{bain2021frozen,xue2022advancing,wang2023internvid,chen2024panda,tan2024vidgen, wang2024koala}, and obtaining high-quality training data is also labor-intensive.

As illustrated in Figure \ref{fig:framework}(b), some controllable I2V generation models adopt well-designed modules for controlling different motion types, such as object movement \cite{wang2024motionctrl,yin2023dragnuwa,wu2025draganything,zhou2024trackgo,zhang2024tora,niu2024mofa,wu2024motionbooth,wang2024objctrl} or camera motion \cite{he2024cameractrl,xu2024cavia,xu2024camco, wang2024cpa}. Some methods are proposed for a single control condition, making it difficult to control multiple motion types simultaneously. Additionally, inputting these control conditions (such as point trajectories or camera extrinsics) may require expert knowledge, which creates a barrier for non-professional users. These methods are less flexible than those that directly control video motion through text input.

To this end, we propose MotionAgent, which enables fine-grained control for text-guided I2V generation. The key design is the motion field agent that converts the motion information in the text prompts into explicit motion representations, providing precise motion guidance. As shown in Figure \ref{fig:framework}(c), our framework directly takes the text as input. First, the agent parses and converts the motion information in the text into object trajectories and camera extrinsics, which explicitly represent object movement and camera motion, respectively. Then, these two intermediate representations are integrated in the 3D space and project into unified optical flow maps through the proposed analytical optical flow composition module. Finally, we tune an optical flow adapter, using the unified flow as conditions, to control a base I2V diffusion model for video generation. Additionally, we design a rethinking mechanism, which forms a feedback loop enabling the agent to correct and improve the former object trajectories and camera extrinsics according to the generated video.

In the experiments, we first evaluate our method on a public I2V generation benchmark \cite{huang2024vbench}. Evaluation results demonstrate that our approach significantly improves the control accuracy of camera motion and achieves comparable video quality with other advanced models. Furthermore, we construct a subset from VBench for assessing the alignment of motion information in the text and the generated video. The results illustrate that our method achieves the best motion accuracy. We also conduct a user study, showing that our generated videos align better with the motion information in the text and maintain high quality. Finally, we conduct ablation studies to verify the effectiveness of the proposed components. And, we demonstrate the robustness and generalization of our method. Our contributions can be summarized as follows:
\begin{itemize}
    \item 
    We propose a novel I2V generation pipeline via a motion field agent, which enables fine-grained motion control for text-guided image-to-video generation.
    \item 
    We construct a new video motion benchmark, a subset of VBench, to assess the alignment of fine-grained motion information in the text and the generated video.
    \item 
    Extensive experiments demonstrate the effectiveness of our proposed method which achieves the most accurate motion control in I2V generation with only text input.
\end{itemize}

\section{Related Work}
\label{sec:relatedwork}

\subsection{Video Generation via Agent}
A series of studies propose to exploit the knowledge of agents for achieving controllable generation \cite{lian2023llm,feng2024layoutgpt,lin2023videodirectorgpt}, zero-shot generation \cite{huang2024free,lu2023flowzero,hong2023direct2v,oh2023mtvg}, or long video generation \cite{zhuang2024vlogger,tian2024videotetris}. Anim-Director \cite{li2024anim} builds an autonomous animation-making agent for generating contextually coherent animation videos. DreamFactory \cite{xie2024dreamfactory} leverages multi-agent collaboration principles and a key frames iteration method to ensure consistency and style across long videos. VIDEOAGENT \cite{soni2024videoagent} aims to improve video plan generation based on a feedback pipeline. ChatCam \cite{liu2024chatcam} introduces a system that navigates camera motion through conversations with users and mimics a professional cinematographer’s workflow. The most similar work is Motionzero \cite{su2023motionzero}, which utilizes LLM to parse the text prompt and warps the intermediate feature maps according to the extracted motion priors to achieve zero-shot video generation.

\subsection{Text-guided I2V Generation}
Generally, most I2V generation models \cite{dai2023animateanything,zhang2024moonshot,ma2024cinemo,jin2024pyramidal} support text and image input simultaneously. The input image guides the visual content of the video, while the text indicates the potential motion. VideoCrafter \cite{chen2023videocrafter1}, Dynamicrafter \cite{xing2025dynamicrafter}, ConsistI2V \cite{ren2024consisti2v}, and I2VGen-XL \cite{zhang2023i2vgen} are all based on the U-Net \cite{ronneberger2015u} backbone and use a text encoder to inject text embeddings into visual features, enabling text control over video content. CogVideoX \cite{yang2024cogvideox} and HunyuanVideo \cite{kong2024hunyuanvideo} based on the DIT \cite{peebles2023scalable} architecture, gain control capability by jointly learning text tokens and visual tokens. These methods, which directly learn the similarity between text embeddings and visual features, often highly depend on the quality of training data. However, there is a lack of large high-quality datasets with text labels that describe video motion. The text labels in most video datasets \cite{bain2021frozen,xue2022advancing,wang2023internvid,chen2024panda,tan2024vidgen,wang2024koala} describe the content of a frame without much motion information and are not suitable for training I2V generation models with strong motion control capability, which leads to poor performance of these methods in achieving precise motion control through direct text input.

\subsection{I2V Generation via Motion Control Module}
With the development of video generation models, controllable video generation gradually attracts more attention. Existing methods \cite{zhang2024tora,xu2024cavia,xu2024camco} propose designing specialized modules for certain motion types, such as object movement or camera motion. DragNUWA \cite{yin2023dragnuwa} encodes sparse point trajectories into dense features as guidance information, which is then injected into the diffusion model to control object motion. DragAnything \cite{wu2025draganything} uses masks to identify a central point and subsequently generates a Gaussian map to track this center, providing a guiding trajectory for object-controllable I2V generation. TrackGo \cite{zhou2024trackgo} introduces a TrackAdapter, which encodes object trajectories into the network through a dual attention mechanism to control object movement in the generated videos. ObjCtrl-2.5D \cite{wang2024objctrl} extends 2D trajectory with depth and uses camera pose to represent the 2.5D object movement. Other video generation methods achieve control of camera motion. MotionCtrl \cite{wang2024motionctrl} directly encodes camera extrinsics as features and inputs them to the model. IMAGE CONDUCTOR \cite{li2024image} and CameraCtrl \cite{he2024cameractrl} both convert sparse camera extrinsics into dense feature maps using Plücker embeddings and employ these dense embeddings to control camera motion. CPA \cite{wang2024cpa} deploy the sparse motion encoding module to transform camera pose into a spatial-temporal embedding for camera control. Most of these methods can only control a single motion type at a time and struggle to achieve simultaneous control of all types.

In addition to encoding a single motion condition into deep features, some methods generate explicit intermediate representations to control multiple motion types of the generated video. Motion-I2V \cite{shi2024motion} generates optical flow maps as the intermediate representation using a flow diffusion model. MOFA-Video \cite{niu2024mofa} proposes a generative motion field adapter, converting sparse point trajectories into dense optical flow through a sparse-to-dense network. AnimateAnything \cite{lei2024animateanything} designs a unified flow generation module, which generates optical flow maps from point trajectories and camera paths. I2VControl \cite{feng2024i2vcontrol} and Motion Prompting \cite{geng2024motion} lift image to 3D and employ point trajectories as explicit intermediate representations for controlling. Diffusion as Shader \cite{gu2025diffusion}, Perception-as-Control \cite{chen2025perception} and MotionCanvas \cite{xing2025motioncanvas} also lift input image into 3D space and utilize different intermediate motion representations for controllable video generation, such as colorize 3D points, UV textures and 3D Bounding-box. Although these models achieve control on various motion types, they often require specific knowledge, such as providing object trajectories or camera extrinsics, which creates a barrier for users.

Different from previous works, our approach proposes a motion field agent that allows direct control of the generated video through text input rather than manual motion representing input. And, we introduce an analytical optical flow composition module, which lifts the image to 3D space and enables our method to simultaneously control both object movement and camera motion.
\section{Method}
\label{sec:method}

\begin{figure*}[t]
\centering
\includegraphics[width=0.99\linewidth]{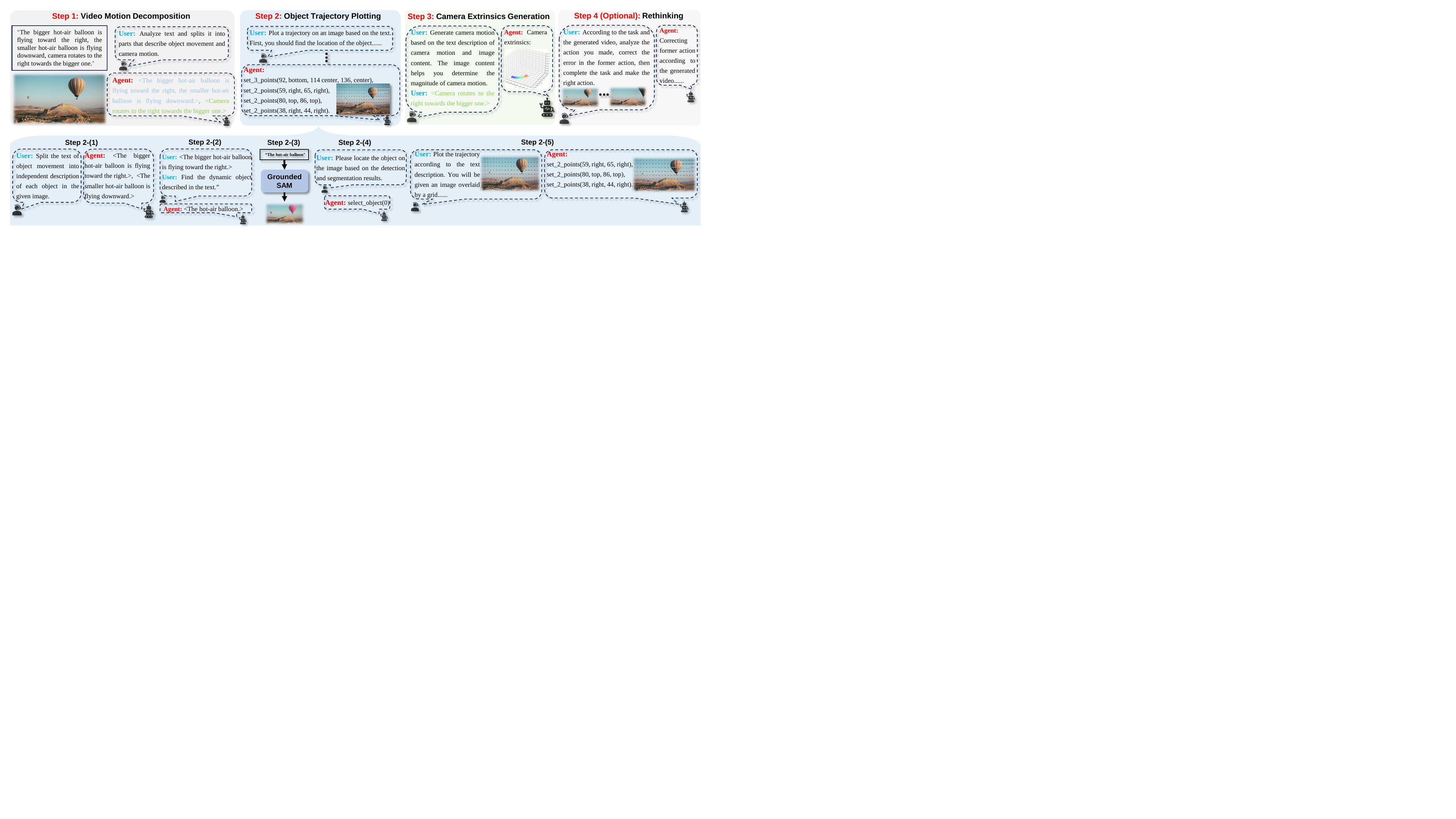}
\caption{Pipeline of Motion Field Agent. \textbf{Step 1:} the agent first parses the input text, dividing the motion information into two parts that respectively describe object movement and camera motion. \textbf{Step 2:} the agent draws the object trajectories according to the text of object movement. \textbf{Step 3:} the agent directly generates the camera extrinsics based on the text of camera motion. \textbf{Step 4 (optional):} the agent rethinks and corrects the former actions according to the generated video.}
\vspace{-0.4cm}
\label{fig:agent}
\end{figure*}

Our method aims to achieve fine-grained controllable I2V generation through text input, allowing precise control of object movement and camera motion. Specifically, a motion field agent first converts the motion information in the text into object trajectories and camera extrinsics. These two explicit intermediate representations are then fed into an improved controllable I2V generation model to guide the video generation. Within the controllable I2V generation model, an analytical optical flow composition module integrates the object trajectories and camera extrinsics to compute unified optical flow maps. Subsequently, we employ Stable Video Diffusion (SVD) \cite{blattmann2023stable} as the base I2V diffusion model and utilize an optical flow adapter \cite{niu2024mofa} as the motion control module to generate the final video.

\subsection{Motion Field Agent}
We generally decompose a video's motion field into object movement within the frame and overall camera motion. Based on this decomposition, our agent performs two key tasks: object trajectory plotting and camera extrinsic generation. We also design a rethinking mechanism, which allows the agent to correct the former actions according to the generated video. The details of the designed agent can be found in Figure \ref{fig:agent}.

\subsubsection{Video Motion Decomposition}
The agent analyzes motion information in the text and splits it into parts that respectively describe object movement and camera motion. This step decouples the complex motion information, allowing for precise and independent control of each motion type.

\subsubsection{Object Trajectory Plotting}
The object trajectory plotting step can be divided into two sub-tasks: object identification and trajectory plotting.

\noindent{\textbf{Object Identification}:
(1) The agent first splits the text of object movement into independent descriptions of each object shown in the given image. (2) Then, the agent further finds the dynamic object described in the text and feeds it into an open-world object detection algorithm. (3) The Grounded-SAM \cite{ren2024grounded} model is called for auxiliary detection and segmentation, facilitating the agent in identifying the corresponding object in the input images. All detection and segmentation results are plotted on the image as semi-transparent masks, which are fed back into the agent for identifying the object.}

\noindent{\textbf{Trajectory Plotting}:
(4) The agent locates the object in the image based on the detection and segmentation results, thereby determining the starting point of the trajectory. (5) The agent plots the trajectory with varying lengths and curvature according to the complexity of the object movement.}

To enhance the capabilities of trajectory plotting, we replace direct trajectory generation with a grid selection approach \cite{zhang2023appagent}. An image is divided into $N \times M$ grids, each labeled by an integer. The agent determines each point of the trajectory by selecting grid numbers, and the trajectory is formed by sequentially connecting all the points. More details of trajectory plotting can be found in the supplementary material.

\subsubsection{Camera Extrinsics Generation}
Agent directly generates camera extrinsics $E$ based on the text of camera motion and input image. The text description specifies the camera's path, and the image helps the agent determine the magnitude of motion. For instance, a wide landscape scene may require intensive camera motion, while a narrow close-up shot may only need slight adjustments of camera location. We constrain the translation $T$ of the camera extrinsics to $(-1, 1)$. In the next optical flow composition module, we rescale the translation $T$ back to a reasonable range based on the estimated depth map.

\subsubsection{Rethinking}
We introduce an optional rethinking mechanism for the agent to correct the former actions based on the already generated results. Specifically, the agent analyzes the generated video and reviews the actions it made at each step previously, which forms a feedback loop and enables the agent to examine and correct the former actions according to the misalignment of text prompt and generated video.

\begin{figure}[t]
\centering
\includegraphics[width=0.99\columnwidth]{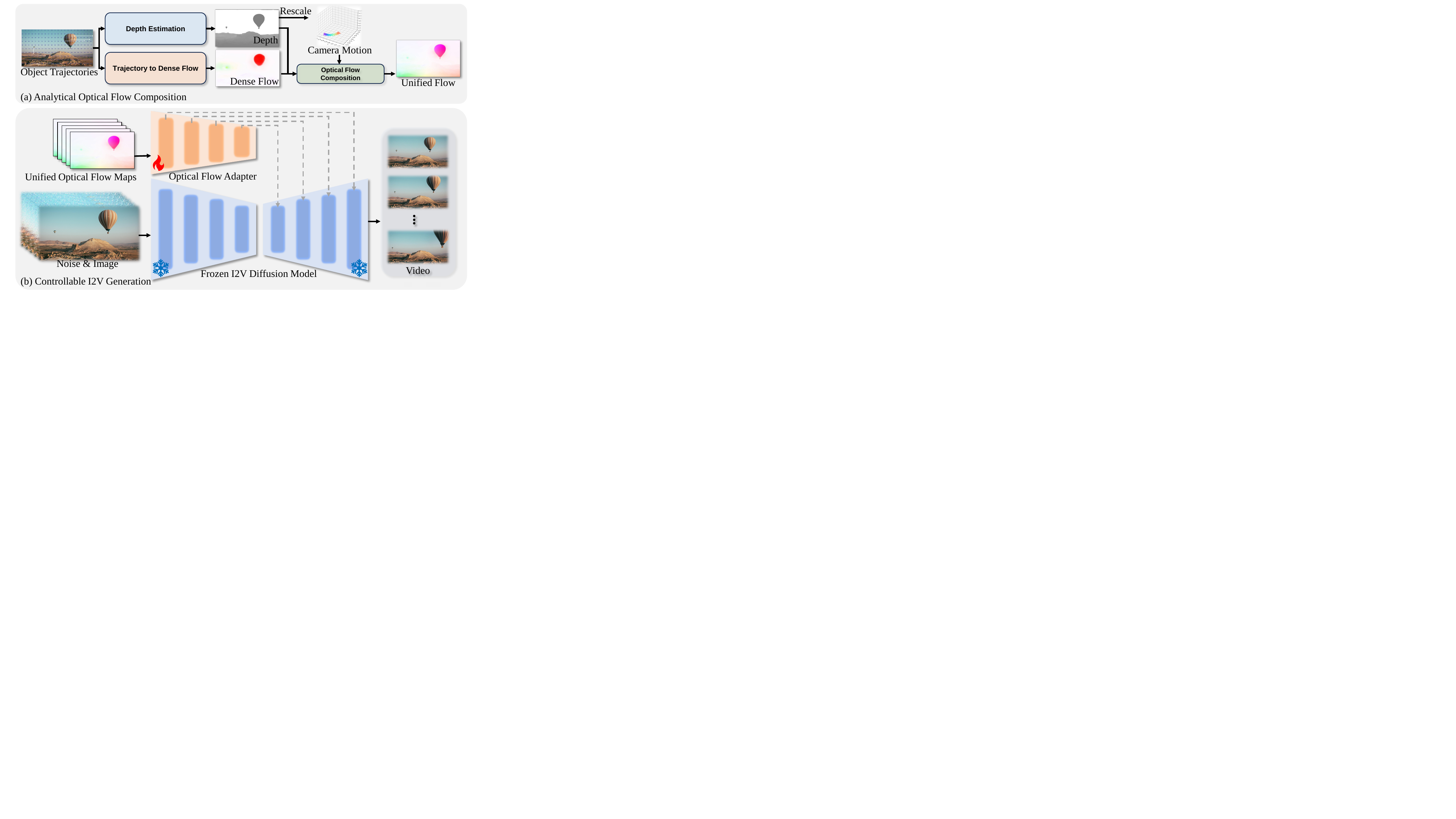}
\caption{Controllable I2V Generation Model. (a) The analytical optical flow composition module calculates unified optical flow maps based on the object trajectories and camera extrinsics. (b) The unified flow maps are fed into a fine-tuned optical flow adapter as the control condition. Then, we generate precisely controlled video results based on a base I2V diffusion model.}
\vspace{-0.4cm}
\label{fig:i2v_model}
\end{figure}

\subsection{Controllable I2V Generation}

After the agent converts the motion information in the text into object trajectories and camera extrinsics, our improved controllable I2V generation model uses these two intermediate representations as inputs, enabling fine-grained control of object movement and camera motion in the video.

\subsubsection{Analytical Optical Flow Composition}
As shown in Figure \ref{fig:i2v_model}(a), this module employs optical flow as a proxy to geometrically compose object movement and camera motion, enabling our I2V generation model to accurately control each motion type in a unified manner.

We first lift the input image to 3D space by using Metric3D \cite{yin2023metric3d} to estimate the depth map $D$ of the input image and unproject each pixel $I^{0}$ in the image to obtain their 3D locations $P^{0}$. CMP \cite{zhan2019self} is utilized to estimate the dense optical flow of object movement $F_{obj}$ from the generated object trajectories. Based on the estimated optical flow map $F_{obj}$ and the depth map $D$, we compute the 3D location offsets $O$ raised by object movement. Then, we add these offsets $O$ to the initial 3D locations $P^{0}$ and get the moved 3D locations ${P}^{1}$.

Subsequently, we reproject these new 3D locations $P^{1}$ into the corresponding image coordinate systems according to the generated camera extrinsics $E$ and calculate the new pixel positions $I^{1}$ in the corresponding image coordinate systems, which can be calculated as,
\begin{equation}
    I^{1} = \Pi(EP^{1}),
\end{equation}
where, $E$ is the generated camera extrinsics, and $\Pi$ is the projection operation. We assume that the camera coordinate of the first frame serves as the world coordinate, and we scale the translation $T$ in the generated camera extrinsics $E$ according to the maximum depth range.

Finally, we compute the offsets of the corresponding pixels in the image coordinate systems as the unified optical flow, which can be noted as,
\begin{equation}
    F = I^{1} - I^{0},
\end{equation}
where $F$ are named unified optical flow maps. These flow maps contain motion information of both object movement and camera motion.

Figure \ref{fig:i2v_model}(b) illustrates the process of controllable I2V generation according to the unified optical flow maps. By adopting the unified optical flow maps as the control conditions, we leverage the optical flow adapter proposed by MOFA-Video \cite{niu2024mofa} and the frozen I2V diffusion model SVD \cite{blattmann2023stable} to achieve controllable I2V generation.

\subsubsection{Optical Flow Adapter Tuning}
Since the unified optical flow maps contain geometrically unrealistic regions compared to the true optical flow maps, directly using these flow maps as the input of the optical flow adapter, which is trained on real optical flow maps, may lead to degraded results in the generated videos. So, we propose fine-tuning the optical flow adapter on the unified optical flow maps.

Specifically, we first use an optical flow model, Unimatch \cite{xu2023unifying}, to estimate the real optical flow. Then we employ an SLAM method, DROID-SLAM \cite{teed2021droid}, to compute the camera extrinsics for each frame. Next, we eliminate the optical flow caused by camera motion based on the estimated camera extrinsics, and approximately obtain the optical flow only caused by object movement. We perform sparse sampling \cite{zhan2019self} on these optical flow maps to obtain sparse object trajectories. More details of data preparation can be found in the supplementary material. Subsequently, we reuse the proposed analytical composition method to calculate the unified optical flow maps utilized as input to fine-tune the optical flow adapter. This step eliminates the domain gap between real and unified flow fields and facilitates high-quality video results.

\section{Experiments}
\label{sec:experiments}
\subsection{Implementation Details}
\label{subsec: implementation}
For the motion field agent, we adopt GPT-4o \cite{achiam2023gpt} as the LLM. We input the image with a resolution of $2560\times1600$ to the agent. In the trajectory plotting module, we divide the image into $20\times10$ grids. For controllable I2V generation, we tune the optical flow adapter module on 32 NVIDIA A800 GPUs. The frozen base I2V generation model is SVD \cite{blattmann2023stable}. We use AdamW as an optimizer. During training, we randomly sample 24 video frames with a stride of 4. The learning rate is set to $2\times10^{-5}$ with a resolution of $512\times512$.

\textbf{Metrics.} In the comparison experiments of the general I2V generation task, we adopt the same evaluation metrics introduced by VBench \cite{huang2024vbench}. In the comparison of controllable I2V generation, we report Object Movement Q\&A, Complex Camera Motion, and Overall Score. All evaluation metrics are higher-the-better. The details of evaluation metrics can be found in the supplementary materials.

\subsection{Comparison with other SOTA Methods}
\subsubsection{General I2V Generation}
\label{subsubsec: general}
\begin{table*}[ht]
  \centering
  \resizebox{1.0\linewidth}{!}
  {
  \begin{tabular}{lccccccccc}
    \toprule
    \multirow{2}{*}{Method} & I2V & Video-Text & Video-Image & Video-Image & Subject & Background & Motion & Aesthetic & Dynamic\\
     & Score & Camera Motion & Subject Consistency & Background Consistency & Consistency & Consistency & Smoothness & Quality & Degree\\
    \midrule
    VideoCrafter \cite{chen2023videocrafter1} & 88.95 & 33.60 & 91.17 & 91.31 & \underline{97.86} & \textbf{98.79} & 98.00 & 60.78 & 22.60 \\
    ConsistI2V \cite{ren2024consisti2v} & 94.81 & \underline{\underline{33.92}} & 95.82 & 95.95 & 95.27 & \underline{98.28} & 97.38 & 59.00 & 18.62 \\
    SEINE \cite{chen2023seine} & 96.26 & 20.97 & 97.15 & 96.94 & 95.28 & 97.12 & 97.12 & 64.55 & \underline{\underline{27.07}} \\
    I2VGen-XL \cite{zhang2023i2vgen} & 96.98 & 13.00 & 97.52 & 97.68 & \underline{\underline{96.36}} & 97.93 & \underline{\underline{98.31}} & \underline{\underline{65.33}} & 24.96 \\
    Animate-Anything \cite{dai2023animateanything} & \textbf{98.31} & 13.08 & \textbf{98.76} & \underline{98.58} & \textbf{98.90} & \underline{\underline{98.19}} & \underline{98.61} & \textbf{67.12} & 2.68 \\
    DynamiCrafter \cite{xing2025dynamicrafter} & \underline{97.98} & \underline{35.81} & \underline{98.17} & \textbf{98.60} & 95.69 & 97.38 & 97.38 & \underline{66.46} & \textbf{47.40} \\
    SVD \cite{blattmann2023stable} & 96.93 & -- & 97.51 & 97.62 & 95.42 & 96.77 & 98.12 & 60.23 & \underline{43.17} \\
    \textbf{MotionAgent} & \underline{\underline{97.51}} & \textbf{81.91} & \underline{\underline{98.06}} & \underline{\underline{98.00}} & 96.10 & 96.76 & \textbf{98.93} & 64.48 & 16.67 \\
    \bottomrule
    \end{tabular}
    }
    \caption{Evaluation results of general I2V generation on VBench \cite{huang2024vbench} (all values are in percentage). The best result is indicated in \textbf{bold}, the second-best result is indicated with \underline{underlines}, and the third-best result is indicated with \underline{\underline{double underlines}}.}
    \vspace{-0.2cm}
    \label{tab:vbench}
\end{table*}

We evaluate the general I2V generation capabilities on a public video generation benchmark, VBench \cite{huang2024vbench}. We compare our model with existing I2V generation models, including VideoCrafter \cite{chen2023videocrafter1}, ConsistI2V \cite{ren2024consisti2v}, SEINE \cite{chen2023seine}, I2VGen-XL \cite{zhang2023i2vgen}, Animate-Anything \cite{dai2023animateanything}, DynamiCrafter \cite{xing2025dynamicrafter}, and our base I2V diffusion model SVD \cite{blattmann2023stable}. To ensure a fair comparison of general I2V generation, we directly use the images and original text prompts provided by VBench as inputs for our model. The evaluation results of other methods are taken directly from the leaderboard on the official website \cite{Vbench}.

As shown in Table \ref{tab:vbench}, our method maintains high-quality I2V generation capabilities. Our pipeline ranks among the top for most metrics and even achieves the best results for some. These results demonstrate that our model can be compatible with general text prompts that may do not include motion information.

Compared with the base model SVD \cite{blattmann2023stable}, our method shows better results in consistency, smoothness, and aesthetic quality metrics. These improvements are attributed to the motion field agent, which can reason and generate suitable object trajectories and camera motion even when there is no motion information included in the text prompts. Additionally, the tuning of the optical flow adapter also improves video quality by constraining the generated video with optical flow and eliminating unreasonable motion. The dynamic degree metric of our method decreases due to the precise control we implement over the generated video. The objects mentioned in the text move accurately, while those not mentioned remain as static as possible, which leads to a lower dynamic degree score. When our model is given a detailed text description with more motion information, it can generate videos with a higher dynamic degree. The qualitative results of dynamic degree metric can be found in the supplementary materials.

Notably, our method achieves $81.91\%$ for the Video-Text Camera Motion metric, which is significantly higher than other methods. This result demonstrates that our method achieves precise control over camera motion according to the text input. The base model SVD does not support text input, while our approach enables SVD to control generated video through text input.

\begin{figure}[t]
\centering
\includegraphics[width=0.99\linewidth]{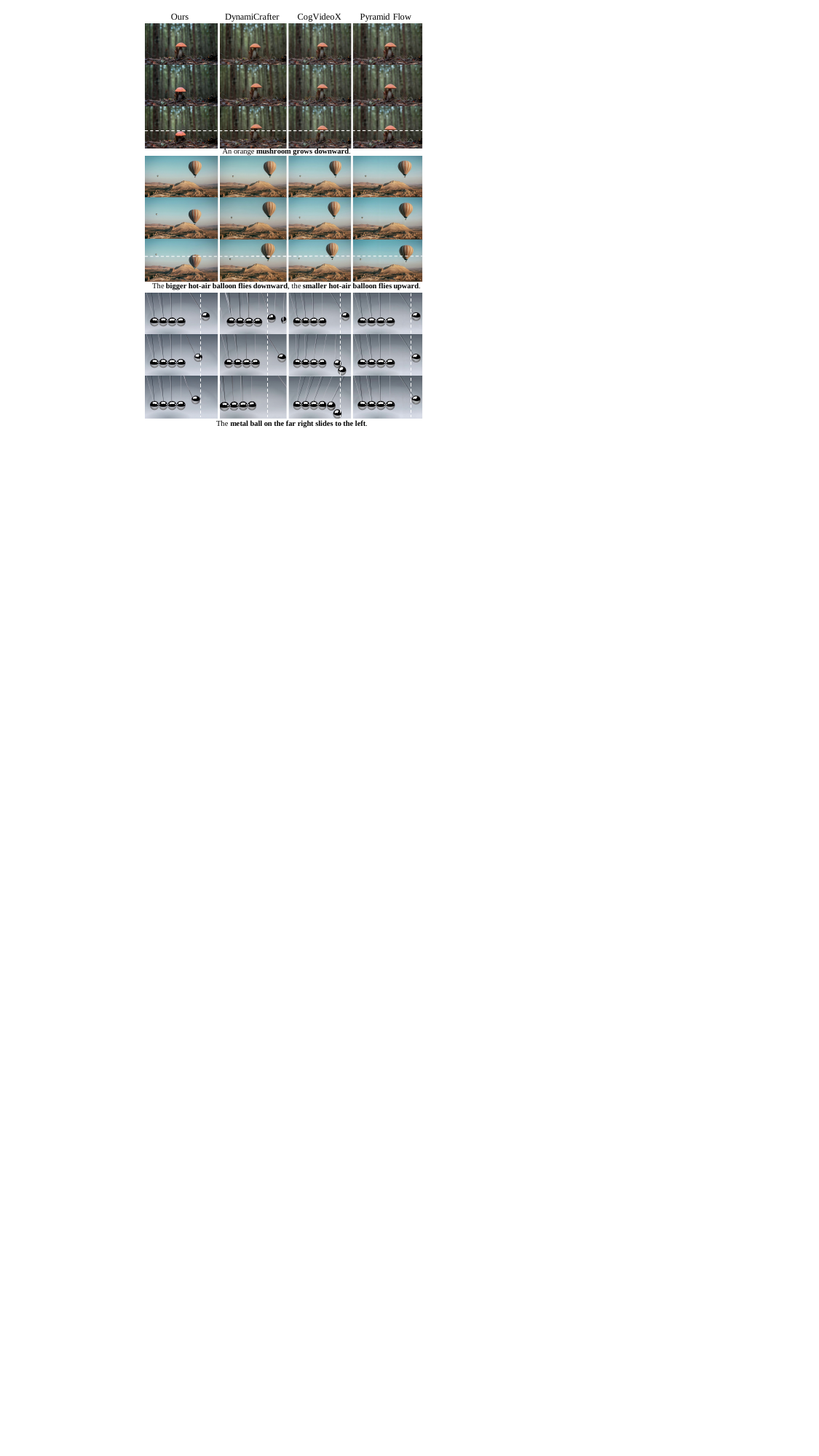}
\caption{Comparison results of controllable I2V generation on our benchmark. The motion described in the text is in \textbf{bold}.}
\vspace{-0.4cm}
\label{fig:comparison}
\end{figure}

\begin{figure*}[t]
\centering
\includegraphics[width=0.99\linewidth]{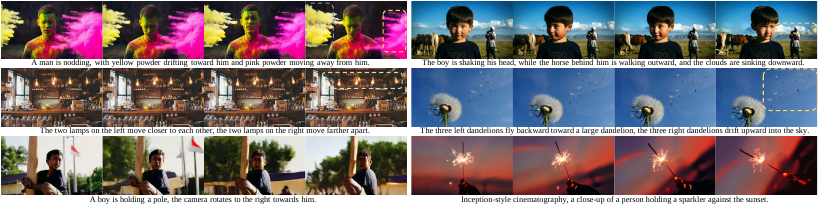}
\vspace{-0.2cm}
\caption{Fine-grained controllable video results generated by our method. Lines 1-2 are multiple object movements control; Line 3 is camera motion control.}
\vspace{-0.4cm}
\label{fig:show}
\end{figure*}

\begin{figure}[t]
\centering
\includegraphics[width=0.80\columnwidth]{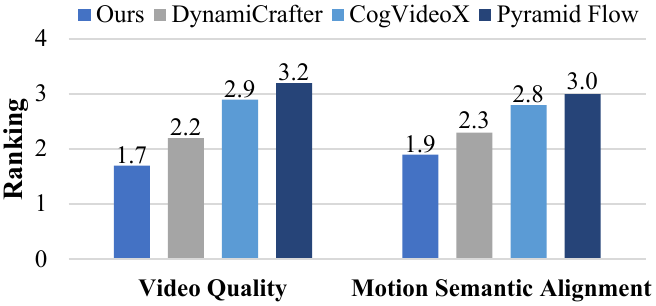}
\vspace{-0.2cm}
\caption{User study results of controllable I2V generation on video quality and semantic alignment of the motion information in text prompt and video.}
\vspace{-0.4cm}
\label{fig:us}
\end{figure}

\subsubsection{Controllable I2V Generation}
\label{subsubsec: controllable}
To further evaluate the control capabilities of I2V generation models through direct text input, we conduct additional assessments. Currently, existing video generation benchmarks do not specifically evaluate the alignment of motion information in the text prompts and the video, so we introduce a new benchmark to evaluate the semantic alignment between videos and input text.

We reuse images from VBench \cite{huang2024vbench} and add more motion information to the original text prompts. To evaluate the control capabilities of object movement, we design 432 new text prompts for 83 images from VBench, providing detailed object movement descriptions. For camera motion, we modify the simple camera motion prompts from 109 images to 662 more complex prompts, such as changing ``zoom in'' to ``first zoom in then zoom out'' or ``zoom in incrementally.''

For each text prompt describing object movement, we design corresponding questions. We utilize a multimodal LLM model, GPT-4o \cite{Chatgpt}, to score the semantic alignment between the object movement in the videos and the input text through a question-and-answer (Q\&A) approach. To evaluate complex camera motions, we adopt the evaluation method introduced by VBench. The compared methods include five UNet-based models: VideoCrafter \cite{chen2023videocrafter1}, ConsistI2V \cite{ren2024consisti2v}, SEINE \cite{chen2023seine}, Motion-I2V \cite{shi2024motion}, and DynamiCrafter \cite{xing2025dynamicrafter}; and two DIT-based models: CogVideoX \cite{yang2024cogvideox} and Pyramid Flow \cite{jin2024pyramidal}.

\begin{table}[t]
  \centering
  \resizebox{0.95\columnwidth}{!} 
  {
  \begin{tabular}{lccc}
    \toprule
    \multirow{2}{*}{Method} & Object Movement & Complex & Total\\
     & Q\&A & Camera Motion & Scores\\
    \midrule
    VideoCrafter \cite{chen2023videocrafter1} & 13.67 & 6.64 & 9.42 \\
    ConsistI2V \cite{ren2024consisti2v} & 24.57 & 6.03 & 13.35\\
    SEINE \cite{chen2023seine} & 21.99 & 1.66 & 9.69\\
    Motion-I2V \cite{shi2024motion} & 28.76 & 7.09 & 15.65\\
    DynamiCrafter \cite{xing2025dynamicrafter} & \underline{\underline{29.38}} & \underline{\underline{8.22}} & \underline{\underline{16.58}}\\
    CogVideoX \cite{yang2024cogvideox} & 26.47 & \underline{20.62} & \underline{22.93}\\
    Pyramid Flow \cite{jin2024pyramidal} & \underline{30.96} & 6.18 & 15.97\\
    \textbf{MotionAgent} & \textbf{45.69} & \textbf{77.76} & \textbf{65.10}\\
    \textbf{MotionAgent} (Rethinking) & 49.58 & 89.04 & 73.45 \\
    \bottomrule
    \end{tabular}
   }
    \caption{Evaluation results of controllable I2V generation on our benchmark (all values are in percentage). The best result (before rethinking) is indicated in \textbf{bold}, the second-best result is indicated with \underline{underlines}, and the third-best result is indicated with \underline{\underline{double underlines}}.}
    \vspace{-0.4cm}
    \label{tab:control}
\end{table}

As shown in Table \ref{tab:control}, our method achieves the best results compared to other methods in controllable video generation tasks. For the Object Movement Q\&A metric, our method demonstrates stronger control over object movement, with higher accuracy scored by GPT-4o \cite{Chatgpt}. It is worth noting that current multimodal LLMs have limited reasoning capabilities, which may lead to some inaccurate judgments and responses. However, the numerical results still indicate that our method achieves the most accurate control over object movement. For the Complex Camera Motion metric, our method significantly outperforms the other methods. Compared to Motion-I2V \cite{shi2024motion}, the optical flow generated by our motion field agent represents the motion information more precisely and results in a higher total score. Moreover, our method maintains consistent performance regardless of whether the camera motion is simple or complex, as illustrated by the camera motion metrics in Tables \ref{tab:vbench} and \ref{tab:control}. Some qualitative comparison results are shown in Figure \ref{fig:comparison}, where our method showcases more precise control over generated videos. 

Our method directly converts motion information in the text prompts into the intermediate motion representations through the motion field agent, and then uses optical flow to achieve controllable video generation. Without the need for high-quality video-image-text pair training data, our approach achieves fine-grained controllable I2V generation. Figure \ref{fig:show} illustrates some Fine-grained controllable video results, and our method achieves precise control over multiple object movements and sophisticated camera motion.

\subsubsection{User Study}
\label{subsubsec: user}
We collect 30 images from VBench \cite{huang2024vbench} and expand the prompts by adding detailed descriptions of object movement and camera motion. We then generate videos using the official codes of DynamiCrafter \cite{xing2025dynamicrafter}, CogVideoX \cite{yang2024cogvideox}, and Pyramid Flow \cite{jin2024pyramidal}. The user study is expected to be completed in 10-20 minutes. For each case, the user study interface shows the videos generated by four methods, and participants are instructed to evaluate the videos from two dimensions: (i) ``Please sort the videos by visual quality from best to worst''; (ii) ``Please sort the videos from high to low based on the semantic alignment of the motion information in the text prompt and the video''. Each case is rated by multiple participants. Finally, we receive 50 valid responses and the final results are calculated based on the 1500 ratings. 

As illustrated in Figure \ref{fig:us}, our method achieves the top rank in both dimensions. These results demonstrate that our method generates high-quality videos with the best alignment of motion information in the text.

\subsection{Ablation Study}
\label{subsec: ablation}
\subsubsection{Object Identification}
We conduct an ablation study on the approach to identify the objects described in the text. In our method, we use Grounded-SAM \cite{ren2024grounded} as an auxiliary detection algorithm to help the agent locate the corresponding objects. Here, we also propose a detection-free approach to achieve object identification through multiple rounds of dialogue.

Specifically, we first ask the agent to determine an initial position, which represents the object described in the text. Then, we plot this initial position directly on the image and input it again along with the text description to the agent. If the agent considers the current initial position correct, it returns the current position directly. If the agent considers the current position incorrect, it returns a new position. The dialogue loop continues until the agent confirms the correctness of the last selected position.

We compare the two approaches for identifying the objects described in the text. As shown in Table \ref{tab:ab}, the results using Grounded-SAM (full model) are significantly better than those using the multi-round dialogue (without a detection tool). The multiround dialogue leads to some incorrect judgments, resulting in a lower score on the Object Movement Q\&A metric. It is worth mentioning that our method decomposes object movement and camera motion, so the degradation caused by incorrect object identification does not significantly affect camera motion in the generated video. Moreover, misjudgments in object identification lead to degradation of the dynamic degree metric.

\begin{table}[t]
  \centering
  \resizebox{0.99\columnwidth}{!} 
  {
  \begin{tabular}{ccccc|ccc}
    \toprule
    Detection & Object & Camera & Flow & Adapter & Object Movement & Complex & Dynamic\\
     Tool & Movement & Motion & Composition & Fine-tuning & Q\&A & Camera Motion & Degree\\
     \midrule
    & \checkmark & \checkmark & \checkmark & \checkmark & 34.33 & 75.11 & 20.53 \\
    \checkmark & & \checkmark & \checkmark & \checkmark & 10.51 & 75.95 & 8.42 \\
    \checkmark & \checkmark & & \checkmark & \checkmark & 38.20 & 0.30 & 29.95 \\
    \checkmark & \checkmark & \checkmark & & \checkmark & 30.07 & 64.92 & 27.89 \\
    \checkmark & \checkmark & \checkmark & \checkmark & & 40.56 & 48.27 & 28.47 \\
    \checkmark & \checkmark & \checkmark & \checkmark & \checkmark & 45.69 & 77.76 & 32.11 \\
    \bottomrule
    \end{tabular}
    }
    \caption{Ablation study of object identification module, optical flow composition module and adapter tuning on our benchmark (all values are in percentage).}
    \vspace{-0.2cm}
    \label{tab:ab}
\end{table}

\begin{figure}[t]
\centering
\includegraphics[width=0.99\columnwidth]{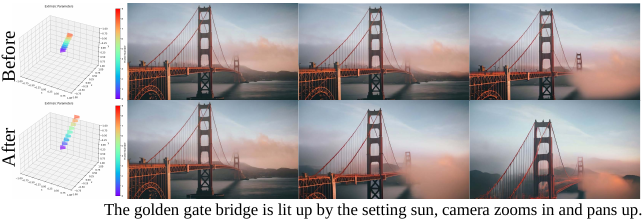}
\caption{Comparison of before and after rethinking process.}
\vspace{-0.4cm}
\label{fig:rethink}
\end{figure}

\subsubsection{Optical Composition and Adapter Fine-tuning}
We first conduct an ablation study on the analytical optical flow composition module and use the original generative motion field adapter proposed by MOFA-Video \cite{niu2024mofa} as a baseline. The baseline can only control object movement or camera motion independently, so we define two models: ``without camera motion'' and ``without object movement'', representing models that can only control object movement or camera motion independently. In our proposed module, we design an analytical approach to compose optical flow. To verify the effectiveness of this approach, we design a mechanism that directly adds the optical flow of object movement and the optical flow caused by camera motion, which we call ``without flow composition''.

As shown in Table \ref{tab:ab}, the baseline models can only control object movement or camera motion independently. Compared to ``without flow composition'', our proposed analytical optical flow composition module achieves higher scores on both the Object Movement Q\&A and Complex Camera Motion metrics. Directly adding optical flows results in many unreasonable areas in the added flow maps, and the object optical flow may be overshadowed by the camera optical flow.

Then, we conduct an ablation study on the effectiveness of optical flow adapter tuning. In Table \ref{tab:ab}, the primitive optical flow adapter is only well suited to optical flow caused by the object movement. When we input unified optical flow maps containing motion information from both object and camera movement, the primitive adapter, without tuning, produces poor results on complex camera motion.

For the dynamic degree metric, our full module achieves the highest scores. Moreover, the results of the dynamic degrees are higher than those acquired from the original prompts in VBench \cite{huang2024vbench}. This showcases that our method generates videos with a higher dynamic degree when there is more detailed motion information in the given text, and follows the text instructions accurately.

\begin{table}[t]
  \centering
  \resizebox{0.99\columnwidth}{!} 
  {
  \begin{tabular}{lccc}
    \toprule
    \multirow{2}{*}{Method} & Object Movement & Complex & Total\\
     & Q\&A & Camera Motion & Scores\\
    \midrule
    MotionAgent(w Qwen2 \cite{bai2023qwen} \& SVD \cite{blattmann2023stable}) & 45.07 & 72.62 & 61.73 \\
    MotionAgent(w Llama3 \cite{dubey2024llama} \& SVD \cite{blattmann2023stable}) & 45.63 & 75.66 & 63.80 \\
    MotionAgent(w GPT-4o \cite{achiam2023gpt} \& Motion-I2V \cite{shi2024motion}) & 44.57 & 71.73 & 61.00 \\
    MotionAgent(w GPT-4o \cite{achiam2023gpt} \& SVD \cite{blattmann2023stable}) & 45.69 & 77.76 & 65.10 \\
    \bottomrule
    \end{tabular}
    }
    \caption{Ablation study results on different LLMs and base I2V generation model on our benchmark (all values are in percentage).}
    \label{tab:ab_1}
    \vspace{-0.4cm}
\end{table}

\subsubsection{Different LLMs and Base I2V Generation model}
As illustrated in Table \ref{tab:ab_1}, we further replace GPT-4o \cite{achiam2023gpt} with other open-sourced LLMs (QWen2 \cite{bai2023qwen} and LLama3 \cite{dubey2024llama}) to verify the robustness of our method. These alternative LLMs achieve almost the same performance, demonstrating that our method is robust with both closed-sourced and open-sourced LLMs. Besides, we also utilize Motion-I2V \cite{shi2024motion} as an additional baseline I2V model to demonstrate the generalization of our method. Specifically, we replace Flow U-Net of Motion-I2V with our method and input the unified optical flow maps into I2V U-Net.

\subsubsection{Rethinking}
In Table \ref{tab:control}, we report the evaluation results before and after the rethinking step on our benchmark. After the rethinking progress, our method improves the metrics of complex camera motion obviously. By analyzing the generated video, our method identifies the camera motion that is not precise enough and generates a more suitable camera path. Moreover, this step also refines the plotted object trajectories, which the metric of object movement can demonstrate. Figure \ref{fig:rethink} shows an example of the rethinking steps, the agent amplifies the translation in the z-axis of the camera motion according to the generated video.
\section{Conclusion}
\label{sec:conclusion}
This work aims to address the challenge of precisely controlling I2V generation using only text input. We first propose a motion field agent that converts the motion information in the text into object trajectories and camera extrinsics. Then, we design an analytical optical flow composition module to integrate the explicit motion representations into unified optical flow maps. Next, we fine-tune an optical flow adapter to achieve controllable I2V generation. Finally, we introduce a rethinking mechanism that enables the agent to correct previous actions according to the generated results. In the experiments, we construct a new video motion benchmark, and the evaluation results demonstrate that our method significantly outperforms other text-guided I2V generation models.
\section{Acknowledgments}
\begin{flushleft}
This research is supported by the MoE AcRF Tier 2 grant (MOE-T2EP20223-0001).
\end{flushleft}
{
    \small
    \bibliographystyle{ieeenat_fullname}
    \bibliography{main}
}
\clearpage
\section{Object Trajectory Plotting Module}
\subsection{Details of Trajectory Plotting}
Inspired by AppAgent \cite{zhang2023appagent}, we replace direct trajectory generation with grid selection based on grid numbers. Specifically, we divide the given image into $N \times M$ grids, breaking it down into small square areas. Each area is labeled with an integer in the top-left corner and subdivided into nine subareas. Based on the previous step, we identify the starting point of the trajectory using the detection result and plot this starting point on the image, represented by a circle. Then, we provide the image overlaid with the grids and starting point as input to the agent.

\begin{figure}[b]
\centering
\includegraphics[width=0.99\columnwidth]{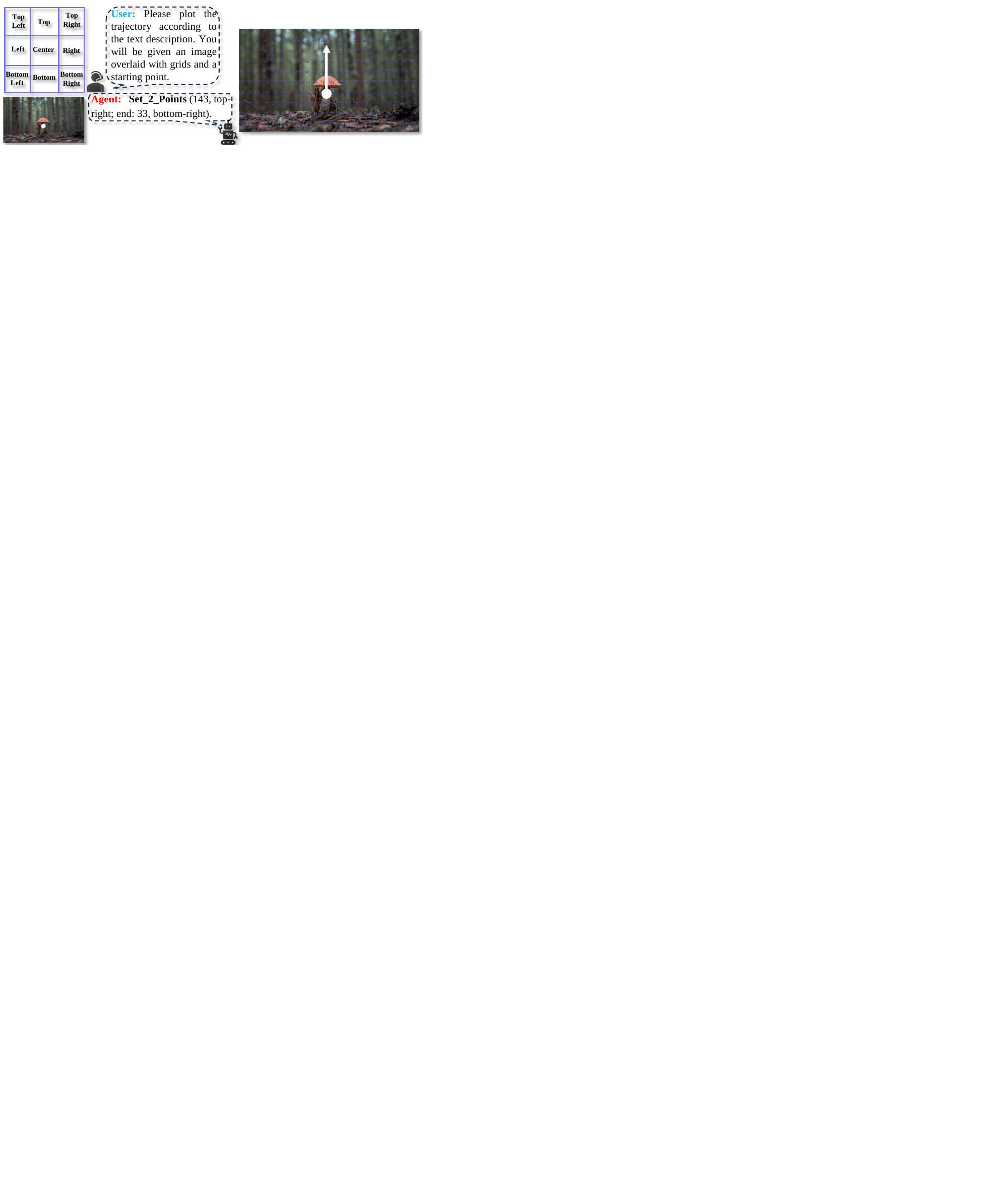}
\caption{Details of trajectory plotting: The grid divides the image into small square areas. Each area is labeled with an integer in the top-left corner and is further subdivided into nine subareas.}
\label{fig:trajectory}
\end{figure}

We define the following functions: \textit{Set\_*\_Points (start: int, string; mid\_*: int, string; end: int, string)} for the agent. Here, \textit{*} is an integer ranging from $1-4$, used to achieve varying lengths and curvature in trajectory plotting. The parameters \textit{start}, \textit{mid\_*}, and \textit{end} include an integer label assigned to the grid area and a string representing the exact location within the grid area. The string can take one of the following nine values: center, top-left, top, top-right, left, right, bottom-left, bottom, and bottom-right.

A simple use case is \textit{Set\_2\_Points (start: 143, top-right; end: 33, bottom-right)}, which sets the starting point of the trajectory at the top-right of grid area 143 and the endpoint at the bottom-right of grid area 33. As illustrated in Figure \ref{fig:trajectory}, this function represents a simple linear trajectory.

Once we obtain the complete object trajectory, we use interpolation to determine the position in the trajectory for each frame. Then, these interpolated positions are input into the subsequent model to calculate dense optical flow.

\label{subsec: ablation}
\begin{table}[h]
  \centering
  \resizebox{0.80\columnwidth}{!} 
  {
  \begin{tabular}{cc|ccc}
    \toprule
    Offset & Grid & Object Movement & Dynamic\\
     Generation & Selection & Q\&A & Degree\\
     \midrule
    \checkmark & & 29.28 & 7.63 \\
    & \checkmark & 45.69 & 32.11 \\
    \bottomrule
    \end{tabular}
    }
    \caption{Ablation study of trajectory plotting module (all values are in percentage).}
    \label{tab:sup_ab}
\end{table}

\subsection{Ablation Study}
Here, we conduct an ablation study on different approaches to trajectory plotting. We use direct offset generation as the baseline module instead of grid selection. Specifically, we provide the agent with the starting point location based on the detection results and ask the agent to directly generate offsets from the starting point to define the object trajectory.

As shown in Table \ref{tab:sup_ab}, the grid selection approach shows better evaluation results in both metrics. The grid selection approach gives the agent an overall understanding of the image layout and is easier for the agent to use than direct offset generation. As for dynamic degree metrics, the offset generation approach cannot output a suitable trajectory length, which may lead to a lower dynamic degree.

\begin{table*}[ht]
  \centering
  \resizebox{1.0\linewidth}{!}
  {
  \begin{tabular}{lccccccccc}
    \toprule
    \multirow{2}{*}{Method} & Video-Text & Video-Image & Video-Image & Subject & Background & Motion & Aesthetic & Dynamic\\
     & Camera Motion & Subject Consistency & Background Consistency & Consistency & Consistency & Smoothness & Quality & Degree\\
    \midrule
    MotionAgent & 81.91 & 98.06 & 98.00 & 96.10 & 96.76 & 98.93 & 64.48 & \textbf{16.67} \\
    MotionAgent (Rethinking) & \textbf{87.02} & \textbf{98.22} & \textbf{98.12} & \textbf{96.44} & \textbf{96.85} & \textbf{99.08} & \textbf{64.69} & 15.38 \\
    \bottomrule
    \end{tabular}
    }
    \caption{Evaluation results of general I2V generation on VBench \cite{huang2024vbench} (all values are in percentage). The best result is indicated in \textbf{bold}.}
    \label{tab:vbench-rethink}
\end{table*}

\begin{figure}[t]
\centering
\includegraphics[width=0.99\columnwidth]{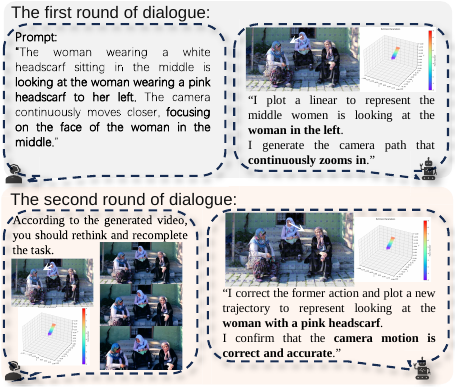}
\caption{Example of rethinking step.}
\label{fig:rethink_sup}
\end{figure}

\section{Rethinking Module}
We report the results on VBench \cite{huang2024vbench} after applying the rethinking step. As shown in Table \ref{tab:vbench-rethink}, the rethinking mechanism achieves higher scores across most metrics. It corrects errors in camera motion generated by earlier stages, leading to notable improvements in the Video-Text Camera Motion metric. Visual quality is also a key consideration in the rethinking step. By refining object movement trajectories and adjusting the range of camera motion, this step effectively reduces artifacts in the generated videos, as reflected in the improved video quality metrics. To reduce such artifacts, the rethinking step tends to shorten object movement trajectories, which may explain the slight decrease observed in the Dynamic Degree metric.

Figure \ref{fig:rethink_sup} shows an example of the rethinking step. The agent corrects the previous error response and confirms the correct action that it made the last time, which facilitates the generation of better video results.

\section{Complex and Ambiguous Prompts.}
We rewrite the motion prompts in our designed sub-VBench using more complex or ambiguous descriptions to evaluate generalization ability and robustness. In Table \ref{tab:control-rethink}, we report comparative results on these rewritten prompts. For complex prompt inputs, our method maintains comparable performance on object movement metrics and shows a slight decline in camera motion compared to the results in Table 2 (Main Body). For ambiguous prompt inputs, performance degradation is observed in both metrics relative to the simple prompts in Table 2 (Main Body). Nevertheless, MotionAgent still achieves competitive results and outperforms other I2V generation methods in achieving semantic alignment between motion prompts and the generated videos.

\begin{table*}[h]
  \centering
  \resizebox{1.0\linewidth}{!}
  {
  \begin{tabular}{lccc|ccc}
    \toprule
    \multirow{3}{*}{Method} & \multicolumn{3}{c|}{Complex Prompts} & \multicolumn{3}{c}{Ambiguous Prompts} \\
    \cmidrule(lr){2-4} \cmidrule(lr){5-7}
    & Object Movement Q\&A & Camera Motion & Total Scores & Object Movement Q\&A & Camera Motion & Total Scores \\
    \midrule
    CogVideoX \cite{yang2024cogvideox} & 24.01 & 16.12 & 19.24 & 20.71 & 11.00 & 14.48 \\
    Pyramid Flow \cite{jin2024pyramidal} & 26.39 & 5.95 & 14.03 & 17.45 & 4.49 & 9.61 \\
    MotionAgent & 45.78 & 73.84 & 62.75 & 37.07 & 66.28 & 54.74 \\
    MotionAgent (Rethinking) & \textbf{48.19} & \textbf{79.32} & \textbf{67.02} & \textbf{42.28} & \textbf{69.37} & \textbf{58.67} \\
    \bottomrule
  \end{tabular}
  }
  \caption{Results of complex and ambiguous controllable I2V generation.}
  \label{tab:control-rethink}
\end{table*}

\section{Training Data Preparation}
To eliminate the domain gap between the unified and real optical flow maps, we propose fine-tuning the optical flow adapter module, which maintains the generation capabilities of the base I2V diffusion model. For each video used for training, we first utilize a binary segmentation model, BiRefNet \cite{zheng2024birefnet}, to decompose the foreground and background. We then remove the dynamic foreground based on the binary segmentation mask. Next, we adopt an SLAM method, DROID-SLAM \cite{teed2021droid}, to compute the camera extrinsics $\hat{E}$ from the masked video and the Metric3D \cite{yin2023metric3d} to estimate the depth map $\hat{D}$ for every frame. Additionally, for the original video, we use an optical flow model, Unimatch \cite{xu2023unifying}, to estimate the real optical flow $\hat{F}$.

Next, we explain how to estimate the optical flow caused by object movement based on the camera extrinsics $\hat{E}$, depth map $\hat{D}$ and real optical flow $\hat{F}$. We define $I^{0}$ as the pixel position in the first frame. According to the predicted real optical flow $\hat{F}$, we compute the corresponding pixel position in the following frames, which can be formulated as,

\begin{equation} 
    I^{1} = I^{0} + \hat{F}. 
\end{equation}

Then, we reproject the pixel position in the following frames back to the image coordinate systems of the first frame according to the depth map $\hat{D}$ and the predicted camera extrinsics $\hat{E}$. This can be computed by,

\begin{equation}
    I^{1}_{obj} = \Pi(\hat{E}^{-1}\Pi^{-1}(I^{1})), 
\end{equation}

where $\Pi$ and $\Pi^{-1}$ are the projection and unprojection operations, respectively. We assume that the camera coordinate of the first frame serves as the world coordinate. $I^{1}_{obj}$ indicates the corresponding pixel position of the following frames in the image coordinate systems of the first frame, which contains only object movement. Finally, we compute the optical flow caused by object movement,

\begin{equation} 
    \hat{F}_{obj} = I^{1}_{obj} - I^{0}. 
\end{equation}

 We perform sampling \cite{zhan2019self} on these optical flow maps of object movement $\hat{F}_{obj}$ to obtain sparse object trajectories. Subsequently, we reuse the proposed analytical composition method to calculate the unified optical flow maps $F$, which are utilized as input to fine-tune the optical flow adapter. Additionally, we calculate the error between the unified optical flow maps $F$ and the real optical flow $\hat{F}$. If the error exceeds a threshold, we replace the unified optical flow maps $F$ with the real optical flow $\hat{F}$ for training.

\section{Metrics Details}
In the comparison experiments of the general I2V generation task, we adopt the same evaluation metrics introduced by VBench \cite{huang2024vbench}. In the comparison of controllable I2V generation, we report Object Movement Q\&A, Complex Camera Motion, and Overall Score.

\textbf{I2V Score} reports the overall score of I2V generation metrics. \textbf{Video-Text Camera Motion} assesses the consistency between camera motion and the input text, such as zoom in/out. \textbf{Video-Image Subject Consistency} assesses whether the appearance of the subject remains consistent throughout the entire video compared to the input image. \textbf{Video-Image Background Consistency} evaluates the temporal consistency of background scenes with the input image. \textbf{Subject Consistency} assesses whether the subject’s appearance remains consistent throughout the entire video. \textbf{Background Consistency} evaluates the temporal consistency of the background scenes across frames. \textbf{Motion Smoothness} evaluates whether the motion in the generated video is smooth and follows the physical laws of the real world. \textbf{Aesthetic Quality} evaluates the artistic and aesthetic value perceived by humans towards each video frame. \textbf{Dynamic Degree} evaluates the level of dynamics generated by each model. \textbf{Object Movement Q\&A} assesses the consistency between the text description and object movement in the video. \textbf{Complex Camera Motion} evaluates the consistency between complex camera movements in the generated video and the input text description. \textbf{Total Score} is the overall metric of controllable I2V generation.

\begin{figure}[b]
\centering
\includegraphics[width=0.99\columnwidth]{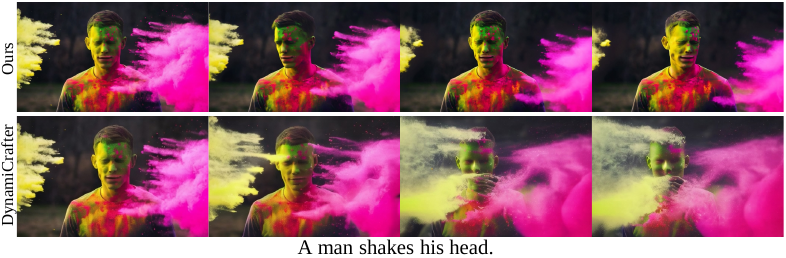}
\caption{Dynamic degree comparison with DynamiCrafter \cite{xing2025dynamicrafter}.}
\label{fig:dynamic}
\end{figure}

\section{Dynamic Degree}
Our method performs a relative lower dynamic degree in the original prompt provided by Vbench \cite{huang2024vbench}, we claim that the decrease is due to the precise control we implement over the generated video. The objects mentioned in the text move accurately, while those not mentioned remain as static as possible, which leads to a lower dynamic degree score. In Figure \ref{fig:dynamic}, we show the result of the comparison with Dynamicrafter \cite{xing2025dynamicrafter} and demonstrate that our method follows the motion information in the text prompt precisely.


\begin{figure}[h]
\centering
\includegraphics[width=0.99\linewidth]{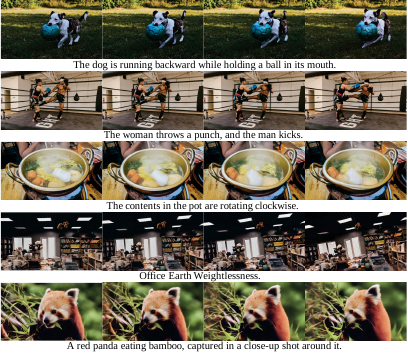}
\caption{More fine-grained controllable video results generated by our method.}
\label{fig:show_sup}
\end{figure}

\begin{figure}[h]
\centering
\includegraphics[width=0.99\linewidth]{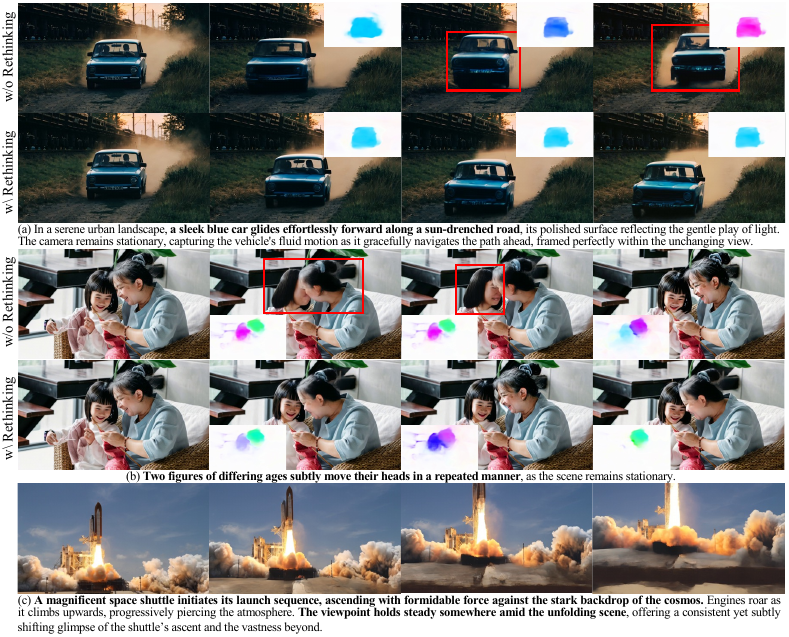}
\caption{Visualization results for rethinking and optical flow.}
\label{fig:comparison-rethink}
\end{figure}

\section{Qualitative Results}

\subsection{Fine-grained Controllable Video Results}
In Figure \ref{fig:show_sup}, we show more fine-grained controllable video results generated by our method.

\subsection{Rethinking Results}
As shown in Figure \ref{fig:comparison-rethink}, the rethinking step corrects the inaccurate object movement (subfigure a, complex prompt) and enhances the quality of the generated video (subfigure b, ambiguous prompt).

\subsection{Visualization of Intermediate Representations}
As illustrated in Figure \ref{fig:comparison-rethink}, we show the intermediate representations generated by the motion field agent.

\subsection{Qualitative Comparison Results}
\begin{figure*}[h]
\centering
\includegraphics[width=0.99\linewidth]{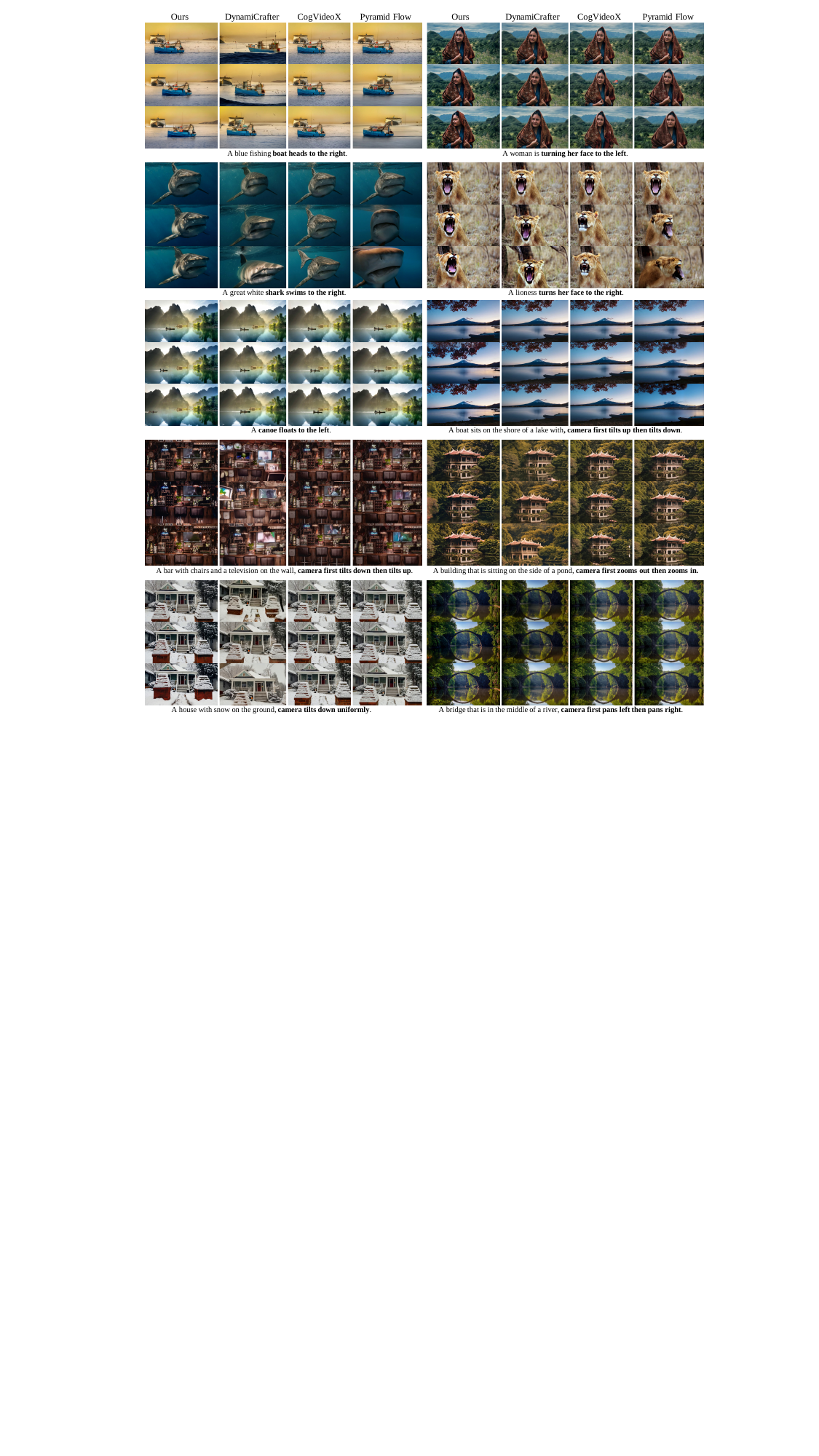}
\caption{More comparison results of controllable I2V generation on our benchmark. The motion described in the text is in \textbf{bold}.}
\label{fig:comparison_sup}
\end{figure*}

In Figure \ref{fig:comparison_sup}, we show more results compared to DynamiCrafter \cite{xing2025dynamicrafter}, CogVideoX \cite{yang2024cogvideox} and Pyramid Flow \cite{jin2024pyramidal}.

\section{User Study Interface}
\begin{figure*}[h]
\centering
\includegraphics[width=0.67\linewidth]{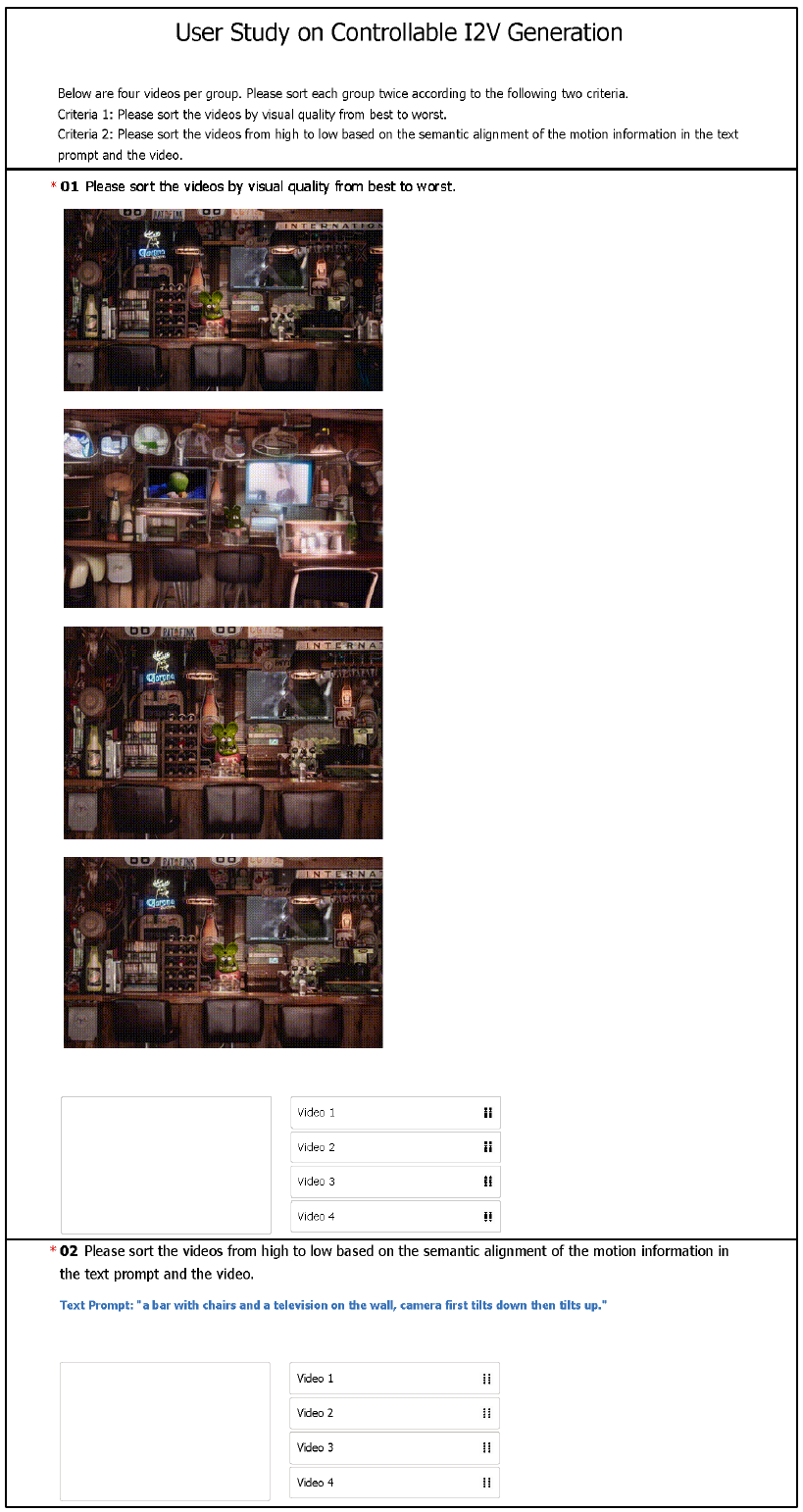}
\caption{User study interface. Each participant is required to evaluate 30 groups of videos and respond to two corresponding
sub-questions for each group. Only one group of videos and two sub-questions are shown here due to the page limit.}
\label{fig:interface}
\end{figure*}

The designed user study interface is shown in Figure \ref{fig:interface}. For each question, we randomly shuffle four videos generated by our method and the other three methods. We then ask participants to rank the videos from highest to lowest twice based on specific requirements. After the user study, we calculate the mean ranking for each method across different evaluation dimensions.

\section{Prompts}
In Table \ref{tab:prompt_object}, \ref{tab:prompt_camera}, \ref{tab:prompt_rethinking}, we present some majority prompts used in our method for object trajectory plotting, camera extrinsics generation, and rethinking.
\clearpage
\begin{table*}[h]
\centering
\begin{tabular}{|p{0.97\linewidth}|}
\hline
\textbf{Prompt Template: Object Trajectory Plotting} \\[0.6em]
\hline
You are an agent trained to plot a trajectory on an image based on a text and a starting point. 
You will be given an image overlaid with a grid and a starting point. 
The grid divides the image into small square areas. 
Each area is labeled with an integer in the top-left corner. 
The starting point of the trajectory is represented by a circle. \\[0.8em]

You can call the following functions to plot a trajectory: \\[0.4em]
-- ...... \\[0.4em]
-- \textbf{Set\_3\_Points(start\_area: int, start\_subarea: str, mid\_area: int, mid\_subarea: str, end\_area: int, end\_subarea: str):}  
This function is used to set a starting point, a mid-point and an end point of a trajectory, which represents a complex trajectory. 
start\_area is the integer label assigned to the grid area, marking the trajectory's starting location. 
start\_subarea is a string representing the exact location to begin the trajectory within the grid area ...... 
The three subareas' parameters can take one of the nine values: center, top-left, top, top-right, left, right, bottom-left, bottom, and bottom-right. \\[0.4em]
-- ...... \\[0.8em]

The task you need to complete is to plot a trajectory to describe: \verb|<task_description>|.  
The location of the trajectory starting point is: \verb|<start_point_location>|.  
Now, given the following labeled image, you need to think and call the function needed to proceed with the task: \\[0.6em]

-- First, find the location of the object according to the task description and set the same starting point as the given one. \\[0.4em]
-- Next, select N midpoints to extend the trajectory. \\[0.4em]
-- Finally, select an end point to complete the trajectory. \\[0.8em]

Your output should include four parts in the given format: \\[0.4em]
-- \textbf{Observation:} Describe what you observe in the image. \\[0.4em]
-- \textbf{Thought:} To complete the given task, what is the step you should take. \\[0.4em]
-- \textbf{Action:} The function call with the correct parameters to proceed with the task. \\[0.4em]
-- \textbf{Summary:} Summarize your actions in one or two sentences. \\[0.6em]
\hline
\end{tabular}
\caption{Prompt template for Object Trajectory Plotting.}
\label{tab:prompt_object}
\end{table*}

\begin{table*}[h]
\centering
\begin{tabular}{|p{0.97\linewidth}|}
\hline
\textbf{Prompt Template: Camera Extrinsics Generation} \\[0.6em]
\hline
You are an agent trained to generate a camera motion based on a text and an image. 
Please note that the world coordinate follows OpenCV convention, where the x-axis points rightwards (camera pans right), the y-axis points downwards (camera tilts down), and the z-axis points frontwards (camera zooms in). \\[0.8em]

You can call the following functions to generate a camera motion: \\[0.4em]
-- ...... \\[0.4em]
-- \textbf{Set\_Camera\_Motion(x\_translation: float, y\_translation: float, z\_translation: float, x\_rotation: int, y\_rotation: int, z\_rotation: int, motion\_type: str):}  
This function sets a simple camera motion, such as pan down, that is represented by the shifting distance and rotation degrees of the camera optical center on the x-axis, y-axis, and z-axis. 
x\_translation is a floating point ranged from (-1.00, 1.00), which represents the shift distance in the x axis ...... 
x\_rotation is an integer ranged from (0, 360), which represents the degrees rotated along the x axis ...... 
motion\_type is a string representing the camera motion type, and the parameters can take one of the three values: uniform, decrement, increment. \\[0.4em]
-- ...... \\[0.8em]

The task you need to complete is to generate a camera motion to describe: \verb|<task_description>|.  
Now, given the following image, you need to think and call the function needed to proceed with the task: \\[0.6em]

-- First, imagine that the given image is shot at the initial location of the camera. \\[0.4em]
-- Then, analyze the text description and the image content to determine the direction and distance of the following camera motion. \\[0.4em]
-- Finally, call the function with the correct parameters to generate the camera motion. \\[0.8em]

Your output should include four parts in the given format: \\[0.4em]
-- \textbf{Observation:} Describe what you observe in the image. \\[0.4em]
-- \textbf{Thought:} To complete the given task, what is the step you should take. \\[0.4em]
-- \textbf{Action:} The function call with the correct parameters to proceed with the task. \\[0.4em]
-- \textbf{Summary:} Summarize your actions in one or two sentences. \\[0.6em]
\hline
\end{tabular}
\caption{Prompt template for Camera Extrinsics Generation.}
\label{tab:prompt_camera}
\end{table*}

\begin{table*}[h]
\centering
\begin{tabular}{|p{0.97\linewidth}|}
\hline
\textbf{Prompt Template: Rethinking} \\[0.6em]
\hline
You are an agent that is trained to rethink and recomplete a specific task about video generation. 
I will provide you some frames of the generated video, which is generated based on the former action you made. 
Additionally, I will describe the task that you should recomplete and give the action you made at the last time. \\[0.8em]

-- According to the task description and the generated video, you should first analyze the former action. \\[0.4em]
-- Then, you should correct the error in the former action. \\[0.4em]
-- Finally, you should recomplete the task and take the right action at this time. \\[0.4em]
-- If you think the action you made last time is correct, you can recomplete the task with the same action. \\[0.8em]

The generated video is: ...... \\[0.6em]
The task you should recomplete is: ...... \\[0.6em]
The action you made last time is: ...... \\[0.6em]
\hline
\end{tabular}
\caption{Prompt template for Rethinking.}
\label{tab:prompt_rethinking}
\end{table*}
\end{document}